\documentclass[10pt,twocolumn,letterpaper]{article}

\usepackage{iccv}
\usepackage{times}
\usepackage{epsfig}
\usepackage{graphicx}
\usepackage{amsmath}
\usepackage{amssymb}

\newcommand{\fig}[1]{Figure~\ref{fig:#1}}

\newcommand{\tab}[1]{Table~\ref{tab:#1}}
\newcommand{\eq}[1]{(\ref{eq:#1})}

\usepackage[breaklinks=true,bookmarks=false]{hyperref}

\iccvfinalcopy 

\def\iccvPaperID{2049} 

\ificcvfinal\fi
\begin{document}

\title{Aggregating Deep Convolutional Features for Image Retrieval}

\author{Artem Babenko\\
Yandex\\
Moscow Institute of Physics and Technology\\
{\tt\small artem.babenko@phystech.edu}
\and
Victor Lempitsky\\
Skolkovo Institute of Science and Technology\\
(Skoltech)\\
{\tt\small lempitsky@skoltech.ru}
}

\maketitle

\begin{abstract}
Several recent works have shown that image descriptors produced by deep convolutional neural networks provide state-of-the-art performance for image classification and retrieval problems. It has also been shown that the activations from the convolutional layers can be interpreted as local features describing particular image regions. These local features can be aggregated using aggregation approaches developed for local features (e.g. Fisher vectors), thus providing new powerful global descriptors.

In this paper we investigate possible ways to aggregate local deep features to produce compact global descriptors for image retrieval. First, we show that deep features and traditional hand-engineered features have quite different distributions of pairwise similarities, hence existing aggregation methods have to be carefully re-evaluated. Such re-evaluation reveals that in contrast to shallow features, the simple aggregation method based on sum pooling provides arguably the best performance for deep convolutional features. This method is efficient, has few parameters, and bears little risk of overfitting when e.g.\ learning the PCA matrix. Overall, the new compact global descriptor improves the state-of-the-art on four common benchmarks considerably.
\end{abstract}

\section{Introduction}

Image descriptors based on the activations within deep convolutional neural networks (CNNs)~\cite{LeCun89} have emerged as state-of-the-art generic descriptors for visual recognition~\cite{Oquab14,Razavian14,Chatfield14}. Several recent works~\cite{Babenko14, Razavian14, Gong14} proposed to use the outputs of last fully-connected network layers as global image descriptors and demonstrate their advantage over prior state-of-the-art when the dimensionality of descriptors is limited.

Recently, research attention shifted from the features extracted from the fully-connected layers to the features from the deep convolutional layers of CNNs~\cite{Cimpoi15, Razavian15, Liu14} (below we refer to these features as \textit{deep convolutional features}). These features possess very useful properties, e.g.\ they can be extracted straightforwardly and efficiently from an image of any size and aspect ratio. Also, features from the convolutional layers have a natural interpretation as descriptors of local image regions corresponding to receptive fields of the particular features. Such features can thus be considered as an analogy of ``shallow'' hand-crafted features such as dense SIFT~\cite{Lowe04,vlfeat}. Perhaps inspired by this analogy, \cite{Long14} suggested to use such features to identify meaningful object parts, while \cite{Cimpoi15} proposed to use Fisher vector \cite{Sanchez13} constructed on these \emph{local} features to produce a \emph{global} image descriptor that provides state-of-the-art classification performance on external datasets.

The focus of this paper is image retrieval and in particular the construction of global descriptors for image retrieval. Following recent papers~\cite{Babenko14, Gong14, Razavian14, Razavian15}, we consider descriptors based on activations of pretrained deep CNNs, and specifically deep convolutional layers of CNNs. Given the emerging perception of the features in the convolutional layers as ``new dense SIFT'' \cite{Long14,Razavian15,Cimpoi15,Liu14}, it seems natural to reuse state-of-the-art embedding-and-aggregation frameworks for dense SIFT such as VLAD \cite{Jegou10}, Fisher vectors \cite{Perronnin10} or triangular embedding \cite{Jegou14}, and apply them to deep convolutional features. Our first contribution is the evaluation of these approaches (specifically, Fisher vectors and triangular embeddings) alongside simpler aggregation schemes such as sum pooling and max pooling. 

Perhaps surprisingly, we have found that the relative performance of the aggregation methods for deep convolutional features is rather different from the case of shallow descriptors. In particular, a simple global descriptor based on sum pooling aggregation without high-dimensional embedding and with simple postprocessing performs remarkably well. Such descriptors based on sum-pooled convolutional features (\textit{SPoC} descriptors) improve considerably the state-of-the-art for compact global descriptors on standard retrieval datasets, and perform much better than deep global descriptors for retrieval  previously suggested in \cite{Babenko14, Gong14, Razavian15}. In addition to the excellent retrieval accuracy, SPoC features are efficient to compute, simple to implement and have almost no hyperparameters to tune. 

Importantly, SPoC features perform better than Fisher vector and triangular embeddings of deep convolutional features. This is in sharp contrast to the dense SIFT case, where sum pooling of raw features does not produce a competitive global descriptor. We further investigate why the performance of deep convolutional features is different from shallow features (SIFT), and show that the preliminary embedding step is not needed for deep convolutional features  because of their higher discriminative ability and different distribution properties. Both qualitative explanation and experimental confirmations for this claim are provided.


Overall, this paper introduces and evaluates a new simple and compact global image descriptor and investigates the reasons underlying its success. The descriptor outperforms the existing methods on the common retrieval benchmarks. For example, the performance of 0.66 mAP on the Oxford dataset with 256-dimensional representation (when using entire images during query process) is achieved. 




\section{Related work}

\textbf{Descriptor aggregation.} The problem of aggregating a set of local descriptors (such as SIFT) into a global one has been studied extensively. The best known approaches are VLAD \cite{Jegou10}, Fisher Vectors \cite{Perronnin10}, and, more recently, triangular embedding \cite{Jegou14}, which constitutes state-of-the-art for ``hand-crafted'' features like SIFT. 

Let us review the ideas behind these schemes (using the notation from \cite{Jegou14}). An image $I$ is represented by a set of features $\{x_1,\dots,x_n\} \subset \mathbf{R}^d$. The goal is to combine these features into a discriminative global representation $\psi(I)$. Discriminativity here means that the representations of two images with the same object or scene are more similar (e.g.\ w.r.t.\ cosine similarity) than the representations of two unrelated images. Apart from discriminativity, most applications have a preference towards more compact global descriptors, which is also a focus of our work here. Consequently, the dimensionality of $\psi(I)$ is reduced by PCA followed by certain normalization procedures.

The common way to produce a representation $\psi(I)$ includes two steps, namely \textit{embedding} and \textit{aggregation} (optionally followed by PCA). The embedding step maps each individual feature $x$ into a higher dimensional vector  $\phi(x) \in \mathbf{R}^D$. Then the aggregation of mapped features $\{\phi(x_1),\dots,\phi(x_n)\} \subset \mathbf{R}^D$ is performed. One possible choice for this step is a simple summation $\psi(I) = \sum \phi(x_i)$ but more advanced methods (e.g.\ \textit{democratic kernel} \cite{Jegou14}) are possible. 

The existing frameworks differ in the choice of the mapping $\phi$. For example, VLAD precomputes a codebook of $K$ centroids $\{c_1,\dots,c_K\}$ and then maps $x$ to vector $\phi_\text{VL}(x) = [0\;0\;\dots,(x-c_k)\;\dots,0] \in \mathbf{R}^{K\times d}$, where $k$ is the number of the closest centroid to $x$. The pipeline for Fisher vector embedding is similar except that it uses the soft probabilistic quantization instead of hard quantization in VLAD. It also includes the second-order information about the residuals of individual features into embedding.
Triangulation Embedding \cite{Jegou14} also uses cluster centroids and embeds an individual feature $x$ by a concatenation of normalized differences between it and cluster centroids $
\phi_\text{TE}(x) = \left [\frac{x - c_1}{||x - c_1||},\dots,\frac{x - c_K}{||x - c_K||} \right ] $.
Then the embeddings $\phi_\text{TE}(x)$ are centered, whitened and normalized.

The rationale behind the embedding step is to improve the discriminative ability of individual features. Without such embedding, a pair of SIFT features $x_i$, $x_j$ coming from unrelated images have a considerable chance of having a large value of the scalar product $\langle x_i,\, x_j\rangle$. This becomes a source of accidental false positive matches between local features, and, if the dataset is big enough, between images (as the similarity between resulting global descriptors is aggregated from similarities between pairs of local features \cite{Bo09,Tolias13}). The embedding methods $\phi(\cdot)$ are typically designed to suppress such false positives. For instance, VLAD embedding suppresses all matches between pairs of features that are adjacent to different centroids in the codebook (making the corresponding scalar product zero). Similar analysis can be performed for other embeddings.

Suppressing false positives with high-dimensional mappings has certain drawbacks. First, such mapping can also suppress true positive matches between local features. Second, the embedding usually includes learning a lot of parameters that can suffer from overfitting if the statistics of training and test sets differ. Likewise, as the representations $\psi(I)$ can be very high-dimensional, it may require hold-out data with similar statistics to learn reliable PCA and whitening matrices. For this reason \cite{Jegou14} proposes to use PCA rotation and power-normalization instead of whitening. Finally, high-dimensional embeddings are computationally intense compared to simpler aggregation schemes. 

Despite these drawbacks, high-dimensional embeddings are invariably used with features like SIFT, since without them the discriminativity of the resulting global descriptors is unacceptingly low. In this paper, we demonstrate that in contrast to SIFT, the similarities of raw deep convolutional features are reliable enough to be used without embedding. Simple sum-pooling aggregation performed on unembedded features thus provides the performance which is comparable with high-dimensional embeddings. Eliminating the embedding step simplifies the descriptor, leads to faster computation, avoids problems with overfitting, and overall leads to a new state-of-the-art compact descriptor for image retrieval.

\begin{figure*}
\centering
\renewcommand{\arraystretch}{0.05}
\begin{tabular}{c}
\includegraphics[width=16.5cm]{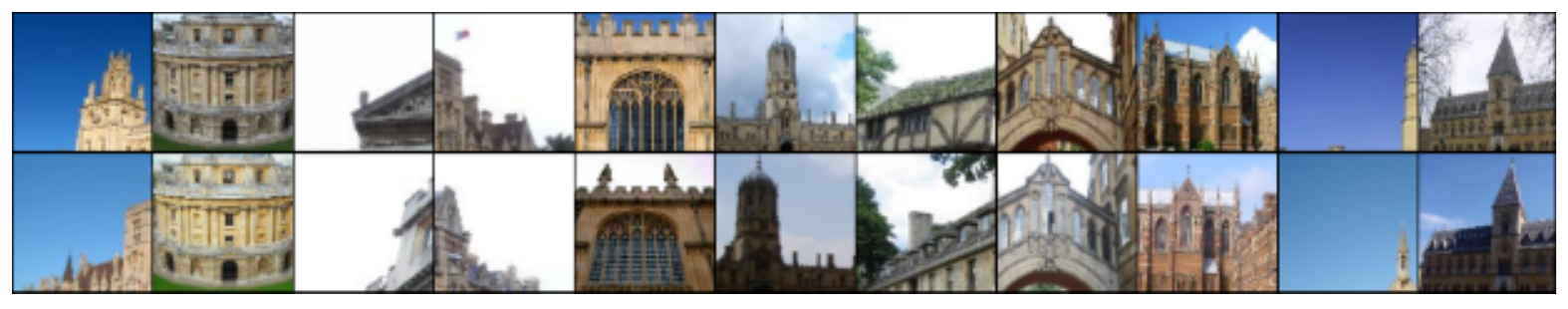}\\
\includegraphics[width=16.5cm]{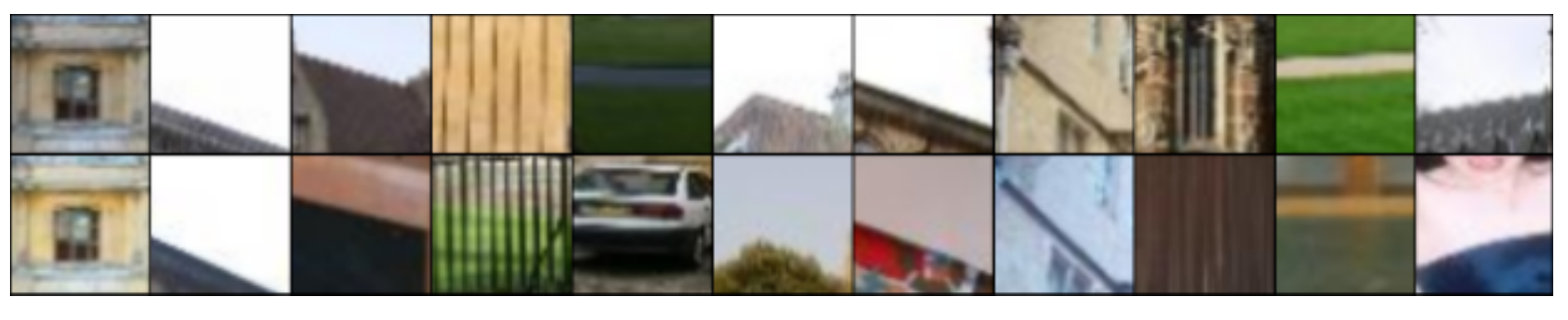}\\
\includegraphics[width=16.5cm]{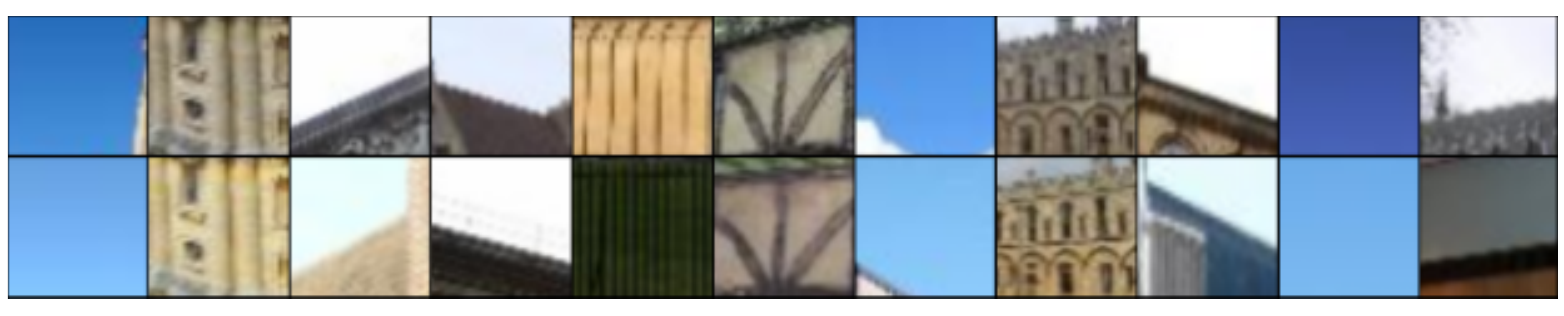}
\end{tabular}
\caption{Randomly selected examples of image patches that are matched by individual deep features (top row), by original SIFT features (middle row) or by Fisher Vector-embedded SIFT features (bottom row). For deep features only the centers of corresponding receptive fields are shown. Overall, the matches produced by deep features has much lower false positive rate.}
\label{fig:matchesComparison}
\end{figure*}

\textbf{Deep descriptors for retrieval.} Several prior works have considered the use of deep features for image retrieval. Thus, the seminal work \cite{Krizhevsky12} have presented qualitative examples of retrieval using deep features extracted from fully-connected layers. After that, \cite{Babenko14} has extensively evaluated the performance of such features with and without fine-tuning on related dataset, and overall reported that PCA-compressed deep features can outperform compact descriptors computed on traditional SIFT-like features. 

Simultaneously, in \cite{Gong14} an even more performant desriptors were suggested based on extracting different fragments of the image, passing them through a CNN and then using VLAD-embedding \cite{Jegou10} to aggregate the activations of a fully-connected layer. Related to that, the work \cite{Razavian14} reported very good retrieval results using sets of few dozen features from fully-connected layers of a CNN, without aggregating them into a global desriptor. 

Finally, the recent works \cite{Azizpour14,Razavian15} evaluated image retrieval descriptors obtained by the max pooling aggregation of the last convolutional layer. Here, we show that using sum pooling to aggregate features on the last convolutional layer leads to much better performance. This is consistent with the interpretation of sum pooling aggregation as an implementation of the simplest match kernel \cite{Bo09}, which is lacking in the case of max pooling.

Overall, compared to previous works \cite{Babenko14,Gong14,Razavian14,Azizpour14,Razavian15} we show that a number of design choices within our descrptor (SPoC) lead to a big boost in descriptor accuracy and efficiency. Compared to those works, we also discuss and analyze the connection to the body of work on descriptor aggregation and evaluate several important aggregation alternatives.

\section{Deep features aggregation}

In this section, we first compare the distribution properties of deep convolutional features and SIFTs and highlight their differences. Based on these differences, we propose a new global image descriptor that avoids the embedding step necessary for SIFTs and discuss several design choices associated with this descriptor. 

In our experiments, deep convolutional features are extracted by passing an image $I$ through a pretrained deep network, and considering the output of the last convolutional layer. Let this layer consist of $C$ feature maps each having height $H$ and width $W$. Then the input image $I$ is represented with a set of $H \times W$ $C$-dimensional vectors, which are the deep convolutional features we work with.

\subsection{Properties of local feature similarities}

As was analysed in e.g.~\cite{Jegou14} individual similarities of raw SIFT features are not reliable, i.e.\ unrelated image patches can result in very close SIFT features. Deep features are expected to be much more powerful as they are learned from massive amount of data in a supervised manner. To confirm this, we have performed a comparison of the properties of similarities computed on the features in the form of two experiments.

{\bf Experiment 1} looks at patches matched by the three types of descriptors (\fig{matchesComparison}). To find these patches we proceed as follows:

\begin{itemize}
  \item For each image in the Oxford Buildings dataset we extract both deep features and dense SIFT features.
  \item We embed SIFT features via Fisher vector embedding with 64 components.
  \item For each feature type (deep convolutional, original SIFT, embedded SIFT), for each query image we compute cosine similarity between its features and the features of all other images in the dataset.
  \item We consider random feature pairs from the top ten list for each image in terms of their similarities and visualize the corresponding image patches (full receptive field for original and embedded SIFT features, the center of the receptive field for deep convolutional features).
\end{itemize}

\fig{matchesComparison} shows the random subset of the feature pair selected with such procedure (one randomly-chosen feature pair per Oxford building), with the top row corresponding to matching based on deep convolutional features, the middle to original dense SIFT, and the bottom to embedded SIFT. As expected, matches produced by deep features have much fewer obvious false positives among them, as they often correspond to the same object with noticeable tolerance to illumination/viewpoint changes and small shifts. SIFT-based matches are significantly worse and many of them correspond to unrelated image patches. The embedding of SIFT features by Fisher vector improves the quality of matches but still performs worse than deep features.

\begin{figure}
\centering
\includegraphics[width=9cm]{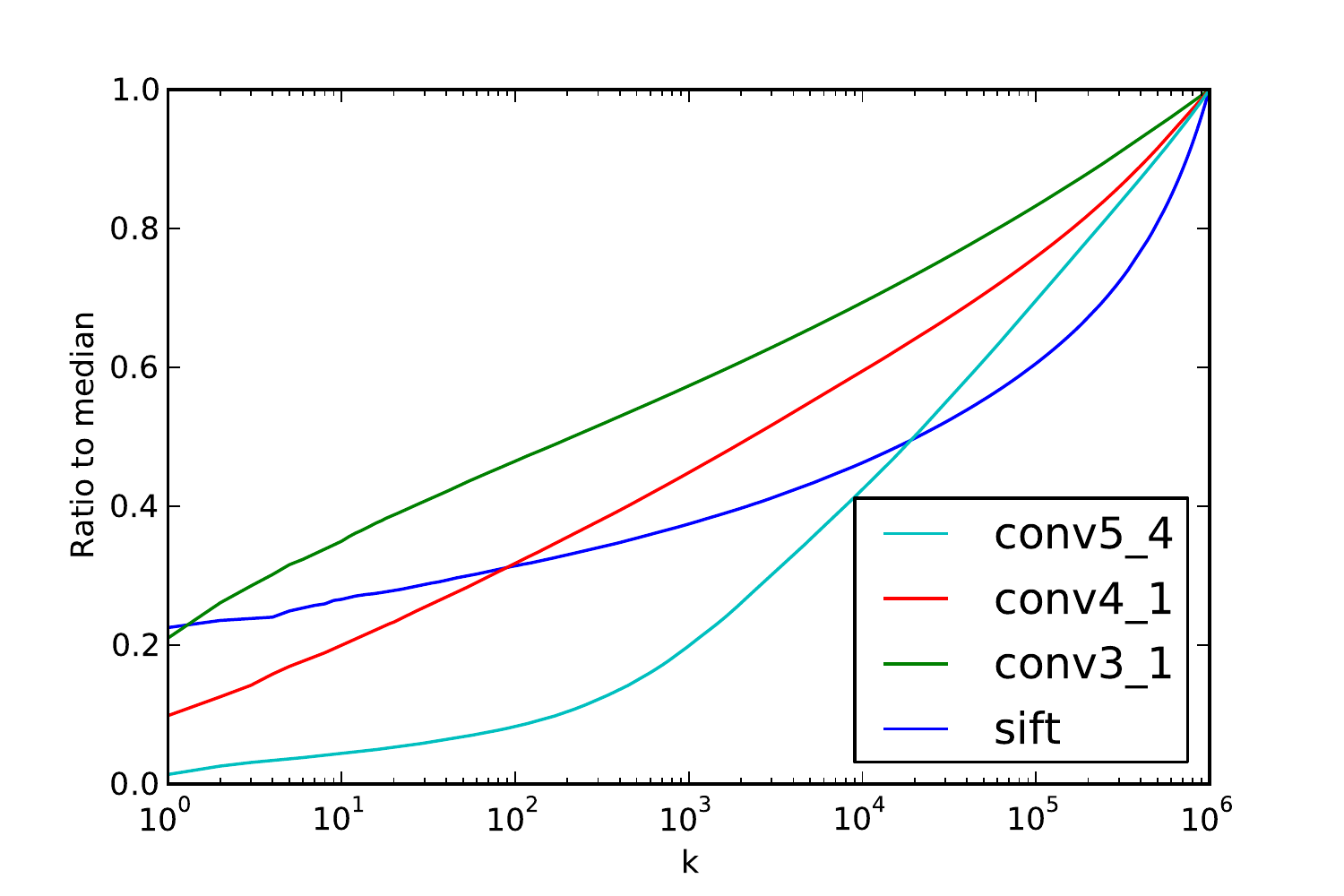}\\
\caption{The average ratio between the distances to the $k$th neighbor and the median distance to all features for dense SIFT and deep convolutional features with the highest norm from three convolutional layers. The features from the last convolutional layer tend to have much closer neighbors (hence much smaller ratios) despite having higher dimensionality thus reflecting the differences in the spatial distribution of the two types of features in the corresponding high-dimensional spaces. }
\label{fig:distanceRatio}
\vspace{-6mm}
\end{figure}

{\bf Experiment 2.} We also investigate the statistics of high-dimensional distributions for deep convolutional features and dense SIFTs. Most of all we are interested in the distribution of deep features with largest norms as these features contribute most to a global descriptor. We also observe them to be the most discriminative by the following experiment. We performed retrieval by sum pooling descriptor but we aggregated only (1) 1\% random features (2) 1\% of features which had the largest norm. The mAP score for the Oxford Buildings dataset \cite{Philbin07} for (1) was only 0.09, which was much smaller than mAP for (2), 0.34. This verifies that features with large norms are much more discriminative than random features.

For different types of features we want to investigate the reliability of matches produced by their individual similarities. To do this, we compare distances from each point to its closest neighbors with  distances to random points in the dataset. In more details, we perform the following. From each query image, we extract ten deep features with maximum norms and for each of them compute the distances to all deep convolutional features of other images. Then we plot a graph which demonstrates how the distance to the $k$-th neighbor depends on its index $k$. For every query feature, distances are normalized by dividing by a median of all distances between the given feature and all features from other images.

We perform this procedure for three types of convolutional features extracted from the layers with different level of depth: "conv3\_1", "conv4\_1" and "conv5\_4" from the OxfordNet~\cite{Simonyan14}. We also perform this experiment for dense SIFTs, though in this case random features from each image were taken as all SIFT features are normalized to have the same norm. For all types of features we use a subset of two million features as the reference set and about a thousand of features per image.

The curves averaged by all queries are shown in \fig{distanceRatio}. They demonstrate that the high-norm deep convolutional features from "conv5\_4" layer have a small amount of ''very close'' neighbors, which are considerably closer than other points. This is in contrast to SIFTs, where typical distances to the closest neighbors are much closer to the distances to random descriptors in the dataset. This fact indicates that closeness of SIFT features is much less informative, and their strong similarities are unreliable and prone to accidental false positive matches. Interestingly, the individual similarities of features from "conv3\_1" and "conv4\_1" are less reliable than from "conv5\_4" (deeper layers produce features with more reliable similarities).

{Note that the second experiment is unsupervised, in the sense that we do not take correctness of matches into account when computing the distances. Rather, the second experiment highlights the substantial differences in the distribution of deep convolutional features and SIFT features in high-dimensional spaces.} 

{The results of both experiments suggest that the individual similarities of deep features from the last convolutional layer are significantly more discriminative and the amount of false positives in matches produced by these similarities should be smaller compared to SIFTs, both because the matching is more accurate (experiment 1) and because higher-norm deep features have fewer close neighbors (experiment 2). This motivates bypassing the high-dimensional embedding step when such features need to be encoded into a global descriptor. }

\subsection{SPoC design}

We describe the \textit{SPoC} descriptor, which is based on the aggregation of raw deep convolutional features without embedding. We associate each deep convolutional feature $f$ computed from image $I$ with the spatial coordinates $(x,y)$ corresponding to the spatial position of this feature in the map stack produced by the last convolutional layer. 

{\bf Sum pooling.} The construction of the SPoC descriptor starts with the sum pooling of the deep features:
\begin{gather}
\psi_1(I) = \sum\limits_{y=1}^{H}\sum\limits_{x=1}^{W}f_{(x,y)} \label{eq:spoc1}
\end{gather}
The scalar product of resulting descriptors corresponds to the simplest match kernel \cite{Bo09} between a pair of images:
\begin{gather}
\text{sim}(I_1,I_2) = \langle \psi(I_1), \psi(I_2) \rangle = \sum\limits_{f_i \in I_1}\sum\limits_{f_j \in I_2}\langle f_i, f_j\rangle
\end{gather}

{\bf Centering prior.} For most retrieval datasets, objects of interest tend to be located close to the geometrical center of an image. SPoC descriptor can be modified to incorporate such centering prior via a simple weighting heuristicы. This heuristics assigns larger weights to the features from the center of the feature map stack, changing the formula \eq{spoc1} to:
\begin{gather}
\psi_2(I) = \sum\limits_{y=1}^{H}\sum\limits_{x=1}^{W}\alpha_{(x,y)}f_{(x,y)}
\end{gather}
Coefficients $\alpha_{(w,h)}$ depend only on the spatial coordinates $h$ and $w$. In particular, we use the Gaussian weighting scheme:
\begin{gather}
\alpha_{(x,y)} = \exp{\left \{-\frac{\left (y - \frac{H}{2}\right )^2+\left (x - \frac{W}{2}\right )^2}{2\sigma^2}\right \}},
\end{gather}
where we set $\sigma$ to be one third of the distance between the center and the closest boundary (the particular choice is motivated from the ''three sigma'' rule of thumb from statistics, although it obviously is not directly related to our use). While very simple, this centering prior provides substantial boost in performance for some datasets as will be shown in the experiments.

{\bf Post-processing.} The obtained representation $\psi(I)$ is subsequently $l_2$-normalized, then PCA compression and whitening are performed:
\begin{equation}
\psi_3(I) = \text{diag}\left(s_1,s_2,\dots,s_N\right)^{-1} M_\text{PCA} \; \psi_2(I)\, \label{eq:spocpca}
\end{equation}
where $M_\text{PCA}$ is the rectangular $N\times C$ PCA-matrix, $N$ is the number of the retained dimensions, and $s_i$ are the associated singular values.

Finally, the whitened vector is $l_2$-normalized:
\begin{equation}
\psi_{SPOC}(I) =  \frac{\psi_3(I)}{\|\psi_3(I)\|_2} \label{eq:spoc}
\end{equation}

Note, that the uncompressed $\psi_2(I)$ has a dimensionality $C$ which equals to the number of output maps in the corresponding convolutional layer. Typical values for $C$ are several hundred hence $\psi(I)$ has moderate dimensionality. Thus, when computing a compact descriptor, it takes much less data to estimate the PCA matrix and associated singular values for SPoC than for Fisher vector or triangulation embedding, since their corresponding descriptors are much higher-dimensional and the risk of overfitting is higher. The experiments below as well as the reports in e.g.~\cite{Jegou14} suggest that such overfitting  can be a serious issue.

\section{Experimental comparison}

\begin{table*}
\centering
\begin{tabular}{|c|c|c|c|c|}
\hline
Method & Holidays & Oxford5K (full) & Oxford105K (full) & UKB\\
\hline
Fisher vector, k=16 & 0.704 & 0.490 & --- & ---\\
\hline
Fisher vector, k=256 & 0.672 & 0.466 & --- & ---\\
\hline
Triangulation embedding, k=1 & 0.775 & 0.539 & --- & ---\\
\hline
Triangulation embedding, k=16 & 0.732 & 0.486 & --- & ---\\
\hline
Max pooling  & 0.711 & 0.524 & 0.522 & 3.57\\
\hline
Sum pooling (SPoC w/o center prior) & 0.802 & 0.589 & 0.578 & 3.65\\
SPoC (with center prior) & 0.784 & 0.657 & 0.642 & 3.66\\
\hline
\end{tabular}
\vspace{1mm}

\caption{ Detailed comparison of feature aggregation methods for deep convolutional features (followed by PCA compression to $256$ dimensions and whitening/normalization).
Sum pooling (SPoC) consistently outperforms other aggregation methods. Full (uncropped) query images are used for Oxford datasets.
\textit{See text for more discussions.}}
\label{tab:mainComparison}
\end{table*}

\begin{figure*}
\centering
\begin{tabular}{c}
\includegraphics[width=16.5cm]{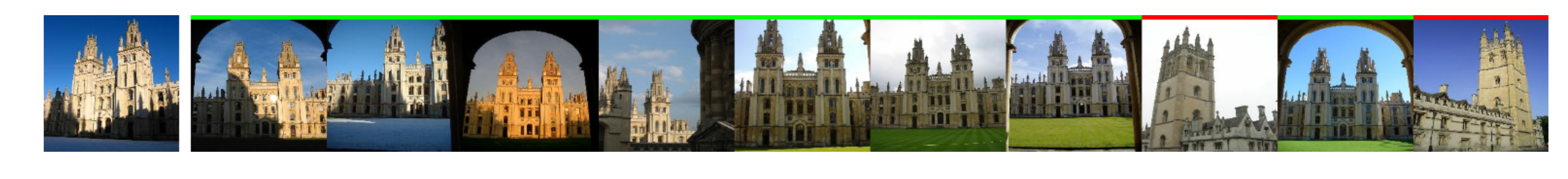}\\
\includegraphics[width=16.5cm]{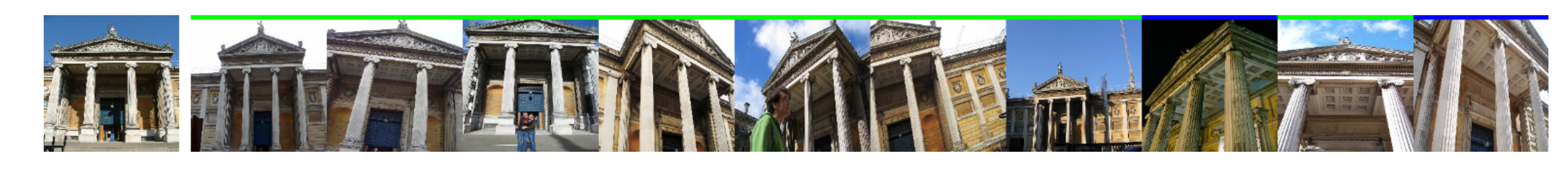}\\
\includegraphics[width=16.5cm]{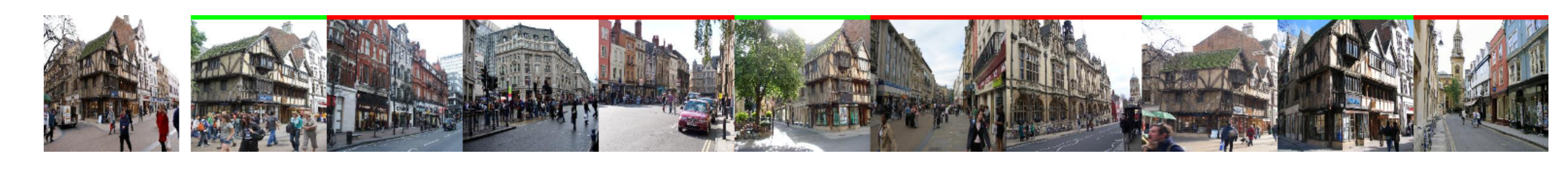}\\
\includegraphics[width=16.5cm]{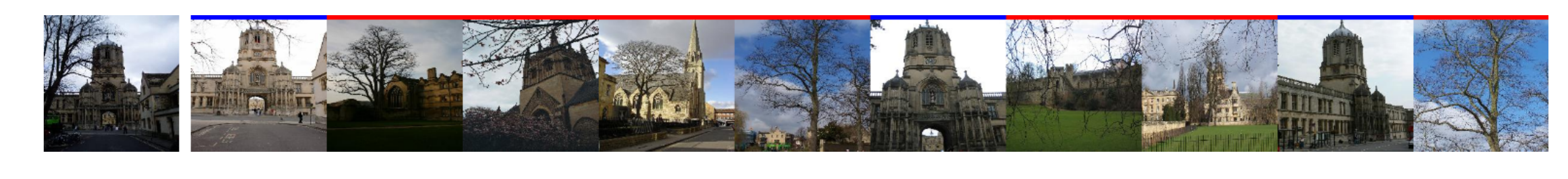}
\end{tabular}
\caption{Retrieval examples (queries and top-ten matches) using SPoC descriptor on the Oxford Buildings dataset (Oxford5K). Red color marks false positives, green color marks true positives and blue color marks images from ''junk'' lists. Two top examples demonstrate that SPoC is robust to changes in viewpoint, cropping and scale. Two bottom rows are the cases where SPoC fails. In these cases SPoC ''is distracted'' by irrelevant objects such as the pavement or the tree.}
\label{fig:examples}
\vspace{-3mm}
\end{figure*}

\begin{figure*}
\centering
\begin{tabular}{c}
\includegraphics[width=16.5cm]{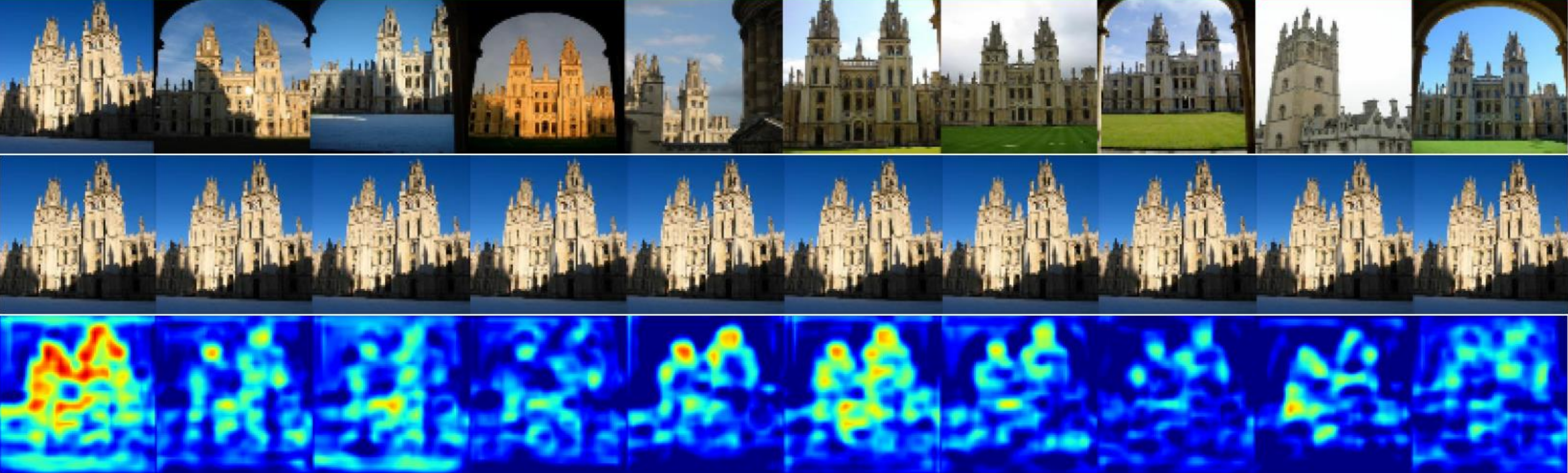}
\end{tabular}
\caption{The examples of similarity maps between the local features of a query image and the SPoC descriptors of top-ten matches. The local features are compressed by the same PCA+whitening matrices as were used for SPoC descriptors and the cosine similarity between each local feature of a query and the SPoC descriptor of a dataset image is computed. The similarity maps allow to localize the regions of a query which are ``responsible'' for the fact that the particular image is considered similar to a query. For instance, for the query above, the spires of the two towers are ``responsible'' for most of the top matches.}
\label{fig:localSimilarities}
\end{figure*}

{\bf Datasets.} We evaluate the performance of SPoC and other aggregation algorithms on four standard datasets.

INRIA Holidays dataset~\cite{Holidays} {\em (Holidays)} contains 1491 vacation snapshots corresponding to 500 groups each having the same scene or object. One image from each group serves as a query. The performance is reported as mean average precision over 500 queries. Similarly to e.g.~\cite{Babenko14}, we manually fix images in the wrong orientation by rotating them by $\pm 90$ degrees.

Oxford Buildings dataset~\cite{Philbin07} {\em (Oxford5K)} contains 5062 photographs from Flickr associated with Oxford landmarks. 55 queries corresponding to 11 buildings/landmarks are fixed, and the ground truth relevance of the remaining dataset w.r.t.\ these 11 classes is provided. The performance is measured using mean average precision (mAP) over the 55 queries.

Oxford Buildings dataset+100K~\cite{Philbin07} {\em (Oxford105K)} contains the Oxford Building dataset and additionally 100K distractor images from Flickr.

University of Kentucky Benchmark dataset~\cite{Nister06} {\em (UKB)} contains $10,200$ indoor photographs of $2550$ objects (four photos per object). Each image is used to query the rest of the dataset. The performance is reported as the average number of same-object images within the top four results.

{\bf Experimental details.} We extract deep convolutional features using the very deep CNN trained by Simonyan and Zisserman~\cite{Simonyan14}. {\tt Caffe}~\cite{Caffe} package for CNNs is used. For this architecture, the number of maps in the last convolutional layer is $C=512$. All images are resized to the size $586\times586$ prior to passing through the network. As a result the spatial size of the last layer is $W\times H=37\times 37$. The final dimensionality for SPoC and, where possible, for other methods is fixed at $N=256$. 

{\bf Aggregation methods.} The emphasis of the experiments is on comparing different aggregation schemes for deep convolutional features.

We consider simple sum pooling and max pooling aggregation. In addition, we consider the two more sophisticated aggregation methods, namely Fisher Vectors~\cite{Perronnin10} (Yael \cite{Yael} implementation) and Triangulation embedding~\cite{Jegou14} (authors implementation). We have carefully tweaked the design choices of these methods in order to adapt them to new kind of features. 

Thus, for Fisher vectors it was found beneficial to PCA-compress the features to 32 dimensions before embedding. For the triangulation embedding several tweaks that had strong impact for SIFTs had relatively small impact in the case of deep features (this includes square rooting of initial features and removing highly-energetic components). We have not used democratic kernel \cite{Jegou14} in the systematic comparisons as it can be applied to all embedding methods, while its computationally complexity can be prohibitive in some scenarios. We observed that for Holidays it consistently improved the performance of triangulation embedding by $~2$ percent (measured prior to PCA).

All embedding methods were followed by PCA reduction to $256$ dimensions. For sum pooling (SPoC) this was followed by whitening, while for Fisher vectors and Triangulation embedding we used power normalization in order to avoid overfitting (as suggested in \cite{Jegou14}). While \cite{Azizpour14} recommends to use whitening with max pooling aggregation, we observed that it reduces retrieval performance and we do not use whitening for max pooling. In the end, all representations were $l2$-normalized and the scalar product similarity (equivalent to Euclidean distance) was used during retrieval. The parameters of PCA (and whitening) were learned on hold-out datasets (Paris buildings for Oxford Buildings, 5000 Flickr images for Holidays) unless noted otherwise.

\begin{table}
\setlength{\tabcolsep}{2pt}
\centering
\begin{tabular}{|c|c|c|}
\hline
Method & Holidays & Oxford5K\\
\hline
Fisher vector, k=16 & 0.704 & 0.490 \\
Fisher vector, PCA on test, k=16 & 0.747 & 0.540\\
\hline
Fisher vector, k=256 & 0.672 & 0.466 \\
Fisher vector, PCA on test, k=256 & 0.761 & 0.581\\
\hline
Triang. embedding, k=1 & 0.775 & 0.539\\
Triang. embedding, PCA on test, k=1 & 0.789 & 0.551\\
\hline
Triang. embedding, k=16 & 0.732 & 0.486\\
Triang. embedding, PCA on test, k=16 & 0.785 & 0.576\\
\hline
Max pooling & 0.711 & 0.524 \\
Max pooling, PCA on test & 0.728 & 0.531 \\
\hline
SPoC w/o center prior & 0.802 & 0.589 \\
SPoC w/o center prior, PCA on test & 0.818 & 0.593 \\
\hline
SPoC (with center prior) & 0.784 & 0.657 \\
SPoC (with center prior), PCA on test & 0.797 & 0.651 \\
\hline
\end{tabular}
\vspace{1mm}

\caption{Comparison of overfitting effect arose from PCA matrix learning for SPoC and other methods. Dimensionalities of all descriptors were reduced to 256 by PCA. Overfitting is much smaller for SPoC and max pooling than for the state-of-the-art high-dimensional aggregation methods.}
\label{tab:overfitting}
\vspace{-2mm}
\end{table}

\begin{table*}
\setlength{\tabcolsep}{4pt}
\centering
\begin{tabular}{|c|c|c|c|c|c|c|c|}
\hline
Method & D & Holidays & \begin{tabular}{@{}c@{}}Oxford5K \\ (full query)\end{tabular} & \begin{tabular}{@{}c@{}}Oxford5K \\ (crop query)\end{tabular} & \begin{tabular}{@{}c@{}}Oxford105K \\ (full query)\end{tabular} & \begin{tabular}{@{}c@{}}Oxford105K \\ (crop query)\end{tabular} &  UKB\\
\hline
SIFT + Triang. + Democr. aggr.\cite{Jegou14} & 1024 & 0.720 & -- & 0.560 & -- & 0.502 & 3.51 \\
SIFT + Triang. + Democr. aggr.\cite{Jegou14} & 128 &  0.617 & -- & 0.433 & -- & 0.353 & 3.40 \\
\hline
Deep fully connected \cite{Babenko14} & 256 & 0.749 & 0.435 & -- & 0.386 & -- & 3.42 \\
Deep fully connected + fine-tuning \cite{Babenko14} & 256 & 0.789 & 0.557 & -- & 0.524 & -- & 3.56 \\
Deep convolutional + Max pooling \cite{Razavian15} & 256 & 0.716 & 0.533 & -- & 0.489 & -- & -- \\
Deep fully connected + VLAD \cite{Gong14} & 512 & 0.783 & -- & -- & -- & -- & -- \\
\hline
Sum pooling (SPoC w/o center prior) & 256 & 0.802 & 0.589 & 0.531 & 0.578 & 0.501 & 3.65\\
\hline
\end{tabular}
\vspace{1mm}

\caption{ Comparison with state-of-the-art for compact global descriptors. For the recent works we report results for dimensionality $256$ or for the closest dimensionalities reported in those papers. Despite their simplicity, SPoC features considerably improve state-of-the-art on all four datasets.}
\vspace{-2mm}
\label{tab:soa}
\end{table*}

{\bf Results.} The comparison of different aggregation methods as well as different variants of SPoC are shown in \tab{mainComparison} and \tab{overfitting}. Several things are worth noting:

\begin{itemize}
\item For deep convolutional features sum pooling emerges as the best aggregation strategy by a margin. It is better than equally simple max pooling, but also better than Fisher vectors and Triangulation embedding even with handicaps discussed below, which is in sharp contrast with SIFT features.

\item We demonstrate the amenability to the overfitting for different methods in \tab{overfitting}. One can see that despite replacing whitening with power normalization, Fisher vectors and Triangulation embedding suffer from the overfitting of the final PCA. When learning PCA on the test dataset their performance improves very considerably. Because of this overfitting effect, it is actually beneficial to use simpler aggregation models: $16$ vs $256$ mixture components for Fisher vectors, $1$ vs $16$ cluster centers in the triangulation embedding. For SPoC and max-pooling overfitting is very small.

\item For triangulation embedding, degenerate configuration with one centroid performs best (more exhaustive search was performed than reported in the table). Even without PCA compression of the final descriptor to 256 dimensions, we observed that the performance of uncompressed descriptor benefitted very little from using more than one centroid, which is consistent with our observations about the statistics of deep convolutional features.


\item Center prior helps for Oxford (a lot), Oxford105K (a lot) and UKB (very little) datasets and hurts (a little) for the Holidays dataset.

\item Whitening is much more beneficial for sum pooling than for max pooling (e.g. max pooling with whitening achieves 0.48 mAP on Oxford while 0.52 without whitening). Apparently some “popular” features that are both common across images and bursty and their contribution to SPoC are suppressed by whitening. For max-pooling burstiness of popular features are less of an issue.

\item PCA compression benefits deep descriptors, as was observed in \cite{Babenko14}. The uncompressed (but still whitened) SPoC features achieve mAP 0.55 on Oxford (0.59 with compression) and 0.796 on Holidays (0.802 with compression).

\end{itemize}

Some qualitative examples of good and bad retrieval examples using SPoC descriptors are shown in \fig{examples}. We also demonstrate some examples of similarity maps between local features of a query image and a global SPoC descriptors of dataset images. To produce these maps we compress the local features by the same PCA+whitening transformation as was used for SPoC construction. Then cosine similarities between local features of the query image and the SPoC descriptor of the dataset image are calculated and visualized as a heatmap. Such heatmaps allow to localize the regions of a query image which are similar to a particular image in the search results. 

{\bf Comparison with state-of-the-art} for compact global descriptors is given in \tab{soa}. Existing works use different evaluation protocols for Oxford datasets, e.g.~\cite{Jegou14,Tolias13} crop query images before retrieval, while recent works \cite{Razavian15,Babenko14,Azizpour14,Razavian14} use uncropped query images. Here, we evaluate our SPoC descriptor in both protocols. In the crop case, for a query image we aggregate only features which have the centers of their receptive fields inside a query bounding box (as it usually done in SIFT-based approaches). As some information about context is discarded by cropping, the results with croped queries are lower.

It turns out that the gap between Oxford5K and Oxford105K performance is quite small for all evaluated settings (especially when queries are not cropped). It seems that the 100K Flickr distractor images while ``distracting enough'' for hand-crafted features, do not really ``distract'' deep convolutional features as they are too different from the Oxford Buildings images.

SPoC features provide considerable improvement over previous state-of-the-art for compact descriptors including deep descriptors in \cite{Babenko14, Gong14, Razavian15}. There are several ways how the results can be further improved. First, a mild boost can be obtained by pooling together features extracted from multiple scales of the same image (about $2$ percent mAP in our preliminary experiments). Similar amount of improvement can be obtained by fine-tuning the original CNN on a specially collected dataset (in the same vein to \cite{Babenko14}).

\section{Summary and Discussion}

We have investigated several alternatives for aggregating deep convolutional features into compact global descriptors, and have suggested a new descriptor (SPoC) based on simple sum-pooling aggregation. While the components of SPoC are simple and well-known, we show that the combination of our design choices results in a descriptor that provides a substantial boost over previous global image descriptors based on deep features and, in fact, over previous state-of-the-art for compact global image descriptors.

Apart from suggesting a concrete descriptor, we have evaluated advanced aggregation strategies proposed for the previous generation of local features (SIFT), and analyzed why sum pooling provides a viable alternative to them for deep convolutional features. In particular, we have highlighted the differences between local convolutional features and dense SIFT. Our experience suggests that deep convolutional features should not be treated as ``new dense SIFT'' in the sense  that the relative performance of different computer vision techniques suggested for features like SIFT has to be reevaluated when switching to new features.

{\small
\bibliographystyle{ieee}
\bibliography{main}

\begin{thebibliography}{10}\itemsep=-1pt

\bibitem{Azizpour14}
H.~Azizpour, A.~S. Razavian, J.~Sullivan, A.~Maki, and S.~Carlsson.
\newblock From generic to specific deep representations for visual recognition.
\newblock {\em CoRR}, abs/1406.5774, 2014.

\bibitem{Babenko14}
A.~Babenko, A.~Slesarev, A.~Chigorin, and V.~S. Lempitsky.
\newblock Neural codes for image retrieval.
\newblock In {\em European Conference on Computer Vision - {ECCV}}, pages
  584--599, 2014.

\bibitem{Bo09}
L.~Bo and C.~Sminchisescu.
\newblock Efficient match kernel between sets of features for visual
  recognition.
\newblock In {\em Advances in Neural Information Processing Systems {(NIPS)}.},
  pages 135--143, 2009.

\bibitem{Chatfield14}
K.~Chatfield, K.~Simonyan, A.~Vedaldi, and A.~Zisserman.
\newblock Return of the devil in the details: Delving deep into convolutional
  nets.
\newblock In {\em British Machine Vision Conference, {BMVC}}, 2014.

\bibitem{Cimpoi15}
M.~Cimpoi, S.~Maji, and A.~Vedaldi.
\newblock Deep filter banks for texture recognition and segmentation.
\newblock In {\em Proceedings of the {IEEE} Conf. on Computer Vision and
  Pattern Recognition ({CVPR})}, 2015.

\bibitem{Yael}
M.~Douze and H.~J{\'{e}}gou.
\newblock The yael library.
\newblock In {\em Proceedings of the {ACM} International Conference on
  Multimedia, {MM}}, pages 687--690, 2014.

\bibitem{Gong14}
Y.~Gong, L.~Wang, R.~Guo, and S.~Lazebnik.
\newblock Multi-scale orderless pooling of deep convolutional activation
  features.
\newblock In {\em 13th European Conference on Computer Vision ({ECCV})}, pages
  392--407, 2014.

\bibitem{Holidays}
H.~J\'egou, M.~Douze, and C.~Schmid.
\newblock Hamming embedding and weak geometric consistency for large scale
  image search.
\newblock In {\em European Conference on Computer Vision}, 2008.

\bibitem{Jegou10}
H.~Jegou, M.~Douze, C.~Schmid, and P.~P{\'{e}}rez.
\newblock Aggregating local descriptors into a compact image representation.
\newblock In {\em The Twenty-Third {IEEE} Conference on Computer Vision and
  Pattern Recognition, {CVPR}}, pages 3304--3311, 2010.

\bibitem{Jegou14}
H.~J{\'{e}}gou and A.~Zisserman.
\newblock Triangulation embedding and democratic aggregation for image search.
\newblock In {\em {IEEE} Conference on Computer Vision and Pattern Recognition,
  {CVPR}}, pages 3310--3317, 2014.

\bibitem{Caffe}
Y.~Jia, E.~Shelhamer, J.~Donahue, S.~Karayev, J.~Long, R.~B. Girshick,
  S.~Guadarrama, and T.~Darrell.
\newblock Caffe: Convolutional architecture for fast feature embedding.
\newblock In {\em Proceedings of the {ACM} International Conference on
  Multimedia, {MM}}, pages 675--678, 2014.

\bibitem{Krizhevsky12}
A.~Krizhevsky, I.~Sutskever, and G.~E. Hinton.
\newblock Imagenet classification with deep convolutional neural networks.
\newblock In {\em Advances in Neural Information Processing Systems {(NIPS)}},
  pages 1106--1114, 2012.

\bibitem{LeCun89}
Y.~LeCun, B.~E. Boser, J.~S. Denker, D.~Henderson, R.~E. Howard, W.~E. Hubbard,
  and L.~D. Jackel.
\newblock Handwritten digit recognition with a back-propagation network.
\newblock In {\em Advances in Neural Information Processing Systems {(NIPS)}},
  pages 396--404, 1989.

\bibitem{Liu14}
L.~Liu, C.~Shen, and A.~van~den Hengel.
\newblock The treasure beneath convolutional layers: Cross-convolutional-layer
  pooling for image classification.
\newblock {\em CoRR}, abs/1411.7466, 2014.

\bibitem{Long14}
J.~Long, N.~Zhang, and T.~Darrell.
\newblock Do convnets learn correspondence?
\newblock In {\em Advances in Neural Information Processing Systems ({NIPS})},
  pages 1601--1609, 2014.

\bibitem{Lowe04}
D.~G. Lowe.
\newblock Distinctive image features from scale-invariant keypoints.
\newblock {\em International Journal of Computer Vision}, 60(2):91--110, 2004.

\bibitem{Nister06}
D.~Nist\'er and H.~Stew\'enius.
\newblock Scalable recognition with a vocabulary tree.
\newblock In {\em IEEE Conference on Computer Vision and Pattern Recognition
  (CVPR)}, 2006.

\bibitem{Oquab14}
M.~Oquab, L.~Bottou, I.~Laptev, and J.~Sivic.
\newblock Learning and transferring mid-level image representations using
  convolutional neural networks.
\newblock In {\em {IEEE} Conference on Computer Vision and Pattern Recognition,
  {CVPR}}, pages 1717--1724, 2014.

\bibitem{Perronnin10}
F.~Perronnin, Y.~Liu, J.~S{\'{a}}nchez, and H.~Poirier.
\newblock Large-scale image retrieval with compressed fisher vectors.
\newblock In {\em The Twenty-Third {IEEE} Conference on Computer Vision and
  Pattern Recognition, {CVPR}}, pages 3384--3391, 2010.

\bibitem{Philbin07}
J.~Philbin, O.~Chum, M.~Isard, J.~Sivic, and A.~Zisserman.
\newblock Object retrieval with large vocabularies and fast spatial matching.
\newblock In {\em {IEEE} Computer Society Conference on Computer Vision and
  Pattern Recognition {(CVPR)}}, 2007.

\bibitem{Razavian14}
A.~S. Razavian, H.~Azizpour, J.~Sullivan, and S.~Carlsson.
\newblock {CNN} features off-the-shelf: An astounding baseline for recognition.
\newblock In {\em {IEEE} Conference on Computer Vision and Pattern Recognition,
  {CVPR} Workshops}, pages 512--519, 2014.

\bibitem{Razavian15}
A.~S. Razavian, J.~Sullivan, A.~Maki, and S.~Carlsson.
\newblock Visual instance retrieval with deep convolutional networks.
\newblock {\em CoRR}, abs/1412.6574, 2014.

\bibitem{Sanchez13}
J.~S{\'{a}}nchez, F.~Perronnin, T.~Mensink, and J.~J. Verbeek.
\newblock Image classification with the fisher vector: Theory and practice.
\newblock {\em International Journal of Computer Vision}, 105(3):222--245,
  2013.

\bibitem{Simonyan14}
K.~Simonyan and A.~Zisserman.
\newblock Very deep convolutional networks for large-scale image recognition.
\newblock {\em CoRR}, abs/1409.1556, 2014.

\bibitem{Tolias13}
G.~Tolias, Y.~S. Avrithis, and H.~J{\'{e}}gou.
\newblock To aggregate or not to aggregate: Selective match kernels for image
  search.
\newblock In {\em {IEEE} International Conference on Computer Vision, {ICCV}},
  pages 1401--1408, 2013.

\bibitem{vlfeat}
A.~Vedaldi and B.~Fulkerson.
\newblock {VLFeat}: An open and portable library of computer vision algorithms.
\newblock \url{http://www.vlfeat.org/}, 2008.

\end{thebibliography}
}

\end{document}